%% file: main.tex
\documentclass[letterpaper, 10 pt, conference]{ieeeconf}  

\IEEEoverridecommandlockouts                              
\usepackage[utf8]{inputenc}
\usepackage[T1]{fontenc}

\usepackage{graphics} 
\usepackage{epsfig} 
\usepackage{amsmath} 
\usepackage{amssymb}  
\usepackage{mathtools}
\usepackage{array}
\usepackage{tabularx}
\newcolumntype{Y}{>{\raggedright\arraybackslash}X}
\usepackage{booktabs}
\usepackage{color}
\let\labelindent\relax
\usepackage{enumitem}
\usepackage{microtype}
\usepackage[hidelinks]{hyperref}

\title{\LARGE \bf
HARP: Handoff-Aware Recovery Planning for Active Target Acquisition in Multi-UAV Handoff
}
\author{
Heegyeong Kim,
Alice James,
Avishkar Seth,
Endrowednes Kuantama,
Richard Han
}

\begin{document}

\maketitle
\thispagestyle{empty}
\pagestyle{empty}

\begin{abstract}

Continuous aerial tracking often requires multiple drones to maintain target coverage over long periods. During handoff, the receiving UAV may lack prior target observations and must actively acquire a usable view without disrupting the sender’s ongoing coverage. This work introduces HARP, a handoff-aware recovery planner that registers bounded sender predictions in the receiver’s local map, distinguishes field-of-view failure from physical occlusion, and generates failure-specific yaw, translation, and parallax viewpoints.  Candidate motions are checked for receiver clearance, sender-fiducial observability, protected sender--target visibility, temporal validity, and deadline compliance. When a complete trajectory cannot be admitted, HARP may execute a sampled admissible nonzero prefix and replan from updated evidence. In simulation, HARP acquired the target in all 49 standard trials and in 15 of 16 stress trials. Field trials validated HARP's recovery geometry under field-of-view limitation and environmental occlusions.

\end{abstract}


\input{Sections/introduction}

\input{Sections/PFandMethod}

\input{Sections/simulation}

\input{Sections/Results_v2}
\input{Sections/Conclusion}


\addtolength{\textheight}{-12cm}   


\bibliographystyle{IEEEtran}
\bibliography{ref.bib}

\end{document}

%% file: Sections/introduction.tex
\section{Introduction}

\begin{figure}
    \centering
    \includegraphics[width=\linewidth]{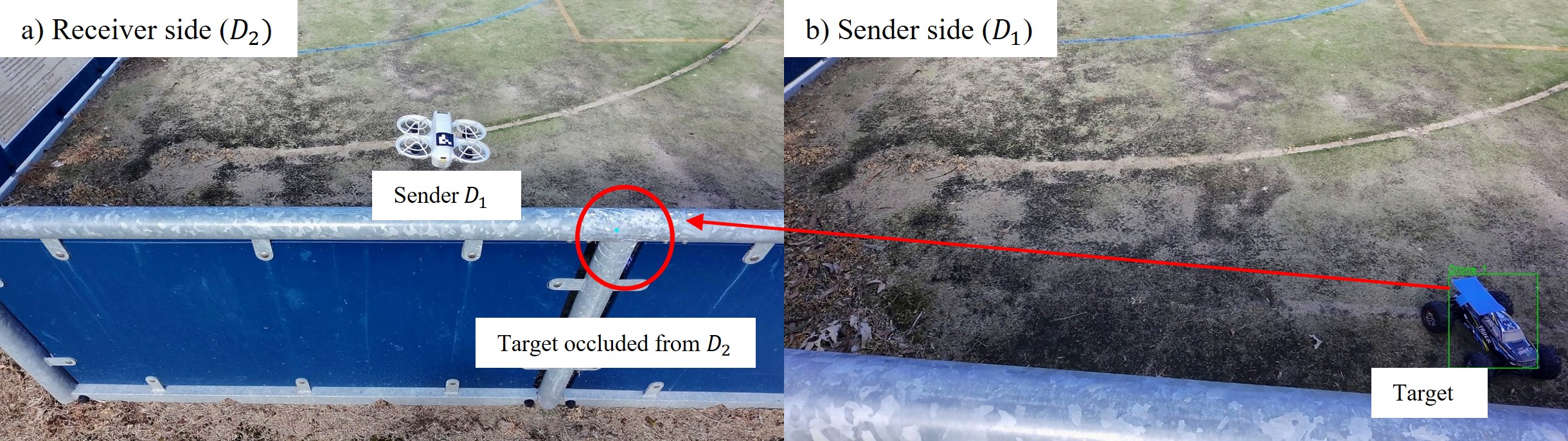}
    \vspace{-15pt}
    \caption{ Representative handoff condition motivating HARP.
(a) The receiver $D_2$ lacks a usable target view because of environmental
occlusion while the sender $D_1$ remains observable.
(b) $D_1$ continues tracking the same target during handoff.
\vspace{-3em}}
    \label{fig:rep}
\end{figure}

Persistent aerial target tracking requires maintaining a usable view as the target moves through the environment~\cite{bhagat2020uav, feng2025deep,sun2024moving}. For long duration monitoring, limited flight endurance creates an additional challenge~\cite{brommer2018long,zhang2025distributed}. Transferring tracking responsibility between UAVs provides a way to extend target observation beyond a single UAV's flight time~\cite{kim2025continuous,leong2021intelligent}. During this handoff, the current tracker $D_1$ can continue to observe the target while the receiver $D_2$ prepares to take over. However, the transfer cannot complete until $D_2$ obtains a usable view of the target from its own viewpoint. This creates a receiver-side acquisition failure. The target may lie outside $D_2$'s Field of View (FoV) or may be physically occluded by the environment, even though $D_1$ continues to track it successfully. In such cases, the sender's target track remains available, but the handoff is stalled because $D_2$ cannot establish the observation required for transfer. Active receiver side visibility recovery can provide the observation needed for handoff while $D_1$ continues tracking the target. Fig.~\ref{fig:rep} illustrates this asymmetric handoff condition in a representative physical setup, where $D_1$ continues tracking the target while the target is occluded from $D_2$'s view point.

This problem differs from conventional visibility-aware aerial tracking. Existing planners maintain visibility of an observed or predicted target while accounting for FoV limits, occlusion, collision avoidance, and target motion~\cite{jeon2019online,wang2021visibility,jeon2020integrated,lee2025qpchaser,hu2025improved}. Recent methods further incorporate active viewpoint adjustment and uncertainty-aware visibility reasoning~\cite{wang2023svpto,gao2024probabilistic}, while interception planners evaluate dynamically feasible motion against predicted target arrival times~\cite{ryou2026Real}. Eva-Tracker addresses reacquisition after temporary visibility loss using the tracking UAV's previous observations~\cite{lin2026evatracker}, while ASPIRe plans informative search trajectories over a maintained target belief when the target is not currently observed~\cite{zhou2024aspire}. Multi-UAV planners coordinate agents to preserve visibility and avoid inter-agent interference~\cite{lee2025dmvc,yin2026visibility}. Perception-side approaches instead address occlusion through robust visual representations and redetection~\cite{wu2025learning,xue2024handling}. Most closely related, Rao et al.~\cite{rao2025multi} use shared teammate visual information for collaborative tracking under occlusion.  In the handoff setting considered here, however, the receiver has not yet observed the target in the current episode and therefore has no local target history on which to base its recovery. It must instead plan its first acquisition using information supplied by the sender, which continues to track the target throughout the handoff interval. This creates an asymmetric planning problem: \textit{one UAV retains a valid target track, whereas the receiver must actively obtain its first observation without disrupting the sender's continuing coverage.} To our knowledge, existing aerial visibility aware tracking, target reacquisition, and multi UAV tracking methods do not directly address this sender-guided first acquisition regime, in which the receiver has no local target observation history and must recover a usable view from time bounded information supplied by the sender. 

We address this problem with HARP, an active visibility recovery framework for visual target handoff between two UAVs. The sender $D_1$ periodically provides a bounded target motion model, which $D_2$ registers into its persistent local map frame and evaluates at candidate arrival times. HARP distinguishes FoV failure from mapped physical occlusion. For FoV recovery, it first searches yaw at the current position and considers translation when no yaw-only endpoint satisfies the configured requirements. For physical occlusion, it uses target--blocker parallax geometry to propose recovery viewpoints. Candidate trajectories undergo sampled checks for receiver clearance, sender-fiducial observability, and interference with the sender--target view, together with motion limits, model validity, and the acquisition deadline. New sender information and local depth observations support revalidation and replanning from the receiver's current state. Actual acquisition is confirmed by a held sequence of fresh receiver detections.


The main contributions of this work are:
\begin{itemize}
\item A motion-planning formulation for sender-guided first acquisition, enabling a receiver UAV with no prior target observations to acquire the target using information from the sending UAV.
\item HARP generates recovery viewpoints for field-of-view and occlusion failures, checks handoff constraints under target uncertainty, and executes sampled-admissible trajectory prefixes when the complete motion artifact cannot be admitted.

\item Validation of HARP through closed-loop simulations, controlled replays, and field trials, demonstrating HARP's recovery behavior under simulated and sensing conditions.

\end{itemize}

%% file: Sections/PFandMethod.tex
\section{Problem Formulation}
\label{sec:problem}

We consider visual handoff of a moving target $T$ from sender UAV $D_1$ to receiver UAV $D_2$. At request time $t_0$, $D_1$ tracks $T$, while $D_2$ has no target observation history in the episode. The objective is sender-guided first acquisition before an externally specified absolute acquisition deadline $T_A$. Continued $D_1$ observation is assumed, while receiver motion is constrained to protect the sender view.

The sender provides associated information
\begin{equation}
\mathcal I_k=(t_k,\mathcal S_{T,k},\mathcal S_{1,k},H_k),
\label{eq:problem_information}
\end{equation}
where $t_k$ is the source time, $\mathcal S_{T,k}$ contains target position and velocity predictions with uncertainty bounds, and $H_k$ is their lifetime. The matched $\mathcal S_{1,k}$ specifies constant sender linear velocity and an intent-validity horizon. Together with the registered source pose, it defines a sender pose prediction with fixed orientation. The transient frame $\mathcal F_v^k$ coincides with $D_1$'s body frame at $t_k$ and remains fixed for message $k$. Receiver navigation and fiducial PnP register it into the persistent map frame $\mathcal F_M$.

Distributed planning uses these registered estimates, receiver navigation, fiducial geometry, and local depth evidence. Target truth and complete scene geometry are excluded from operational inputs. Let $\mathbf q_i(t)\in SE(3)$ denote UAV body poses in $\mathcal F_M$. The receiver maintains
\begin{equation}
\mathcal M_2(t)=(\mathcal O_2(t),\mathcal K_2(t),\mathcal U_2(t)),
\end{equation}
representing OCCUPIED, known-FREE, and UNKNOWN space. The map rolls within the fixed frame $\mathcal F_M$, treating space outside its coverage as UNKNOWN. UNKNOWN cannot certify free clearance or map-supported visibility.

For query time $\tau$,
\begin{equation}
\Delta_k(\tau)=\tau-t_k,\qquad
0\leq\Delta_k(\tau)\leq H_k.
\label{eq:problem_model_time}
\end{equation}
The support $\mathcal X_T^k(\tau)\subset\mathbb R^3$ describes target-centroid uncertainty, excluding physical target extent. It is propagated in $\mathcal F_v^k$ and rigidly registered without re-axis-alignment. Related reachable-area reasoning has been used to account for target-prediction uncertainty in aerial tracking~\cite{lee2025qpchaser}.

Let $\Pi_2$, $z_2$, and $\Omega_2^\delta$ denote calibrated target-camera projection, camera depth, and the usable image region including image and detector margins. Projection requires positive depth and finite image coordinates. Define
\begin{equation}
\begin{aligned}
\mathsf{FoVReady}(\mathbf q_2,\tau)
\Longleftrightarrow\;&
\forall\mathbf p\in\mathcal X_T^k(\tau):\\
&z_2(\mathbf p;\mathbf q_2)>0
\land \Pi_2(\mathbf p;\mathbf q_2)\in\Omega_2^\delta .
\end{aligned}
\label{eq:problem_fov}
\end{equation}
At replanning time $t_r$, a receiver trajectory satisfies
\begin{equation}
\boldsymbol\xi_2:[0,T_\xi]\rightarrow SE(3),
\qquad \boldsymbol\xi_2(0)=\mathbf q_2(t_r).
\end{equation}
FoV recovery requires endpoint image readiness. OCC recovery additionally requires complete-support visibility through known-FREE viewing segments, excluding target endpoints. During motion, the receiver clearance envelope must remain known-FREE, the sender fiducial must remain observable, and the receiver envelope must avoid the protected sender--target viewing volume. Motion limits, model and sender validity, and the deadline also constrain admissibility. The protected-view constraint addresses inter-agent occlusion, which has also been studied in multi-UAV tracking~\cite{lee2025dmvc}. Sections~\ref{sec:method_certification} and~\ref{sec:method_execution} describe the sampled checks and their use during execution.

Let $\mathsf{Acq}_2(t)$ require a fresh, held sequence of actual target observations, with thresholds specified in Simulation Experiments. Define
\begin{equation}
t_{\mathrm{acq}}=\inf\{t\geq t_0:\mathsf{Acq}_2(t)=1\},
\label{eq:problem_acquisition}
\end{equation}
where $\inf\varnothing=+\infty$. Recovery succeeds if $t_{\mathrm{acq}}\leq T_A$. Acquisition may occur during motion before geometric arrival. PATH identity verification and responsibility transfer lie outside the acquisition task evaluated here.

The operating model assumes calibrated camera/body geometry, timestamp-consistent registration and navigation, target-centroid containment within the declared support, and constant sender velocity with fixed orientation during its validity interval. 
The sampled admissibility checks are conditional on these assumptions and do not provide continuous-time safety or robustness guarantees under arbitrary motion, localization or mapping errors, or dynamic obstacles outside this model. 

\section{HARP Method}
\label{sec:method}

\begin{figure}
    \centering
    \includegraphics[width=\linewidth]{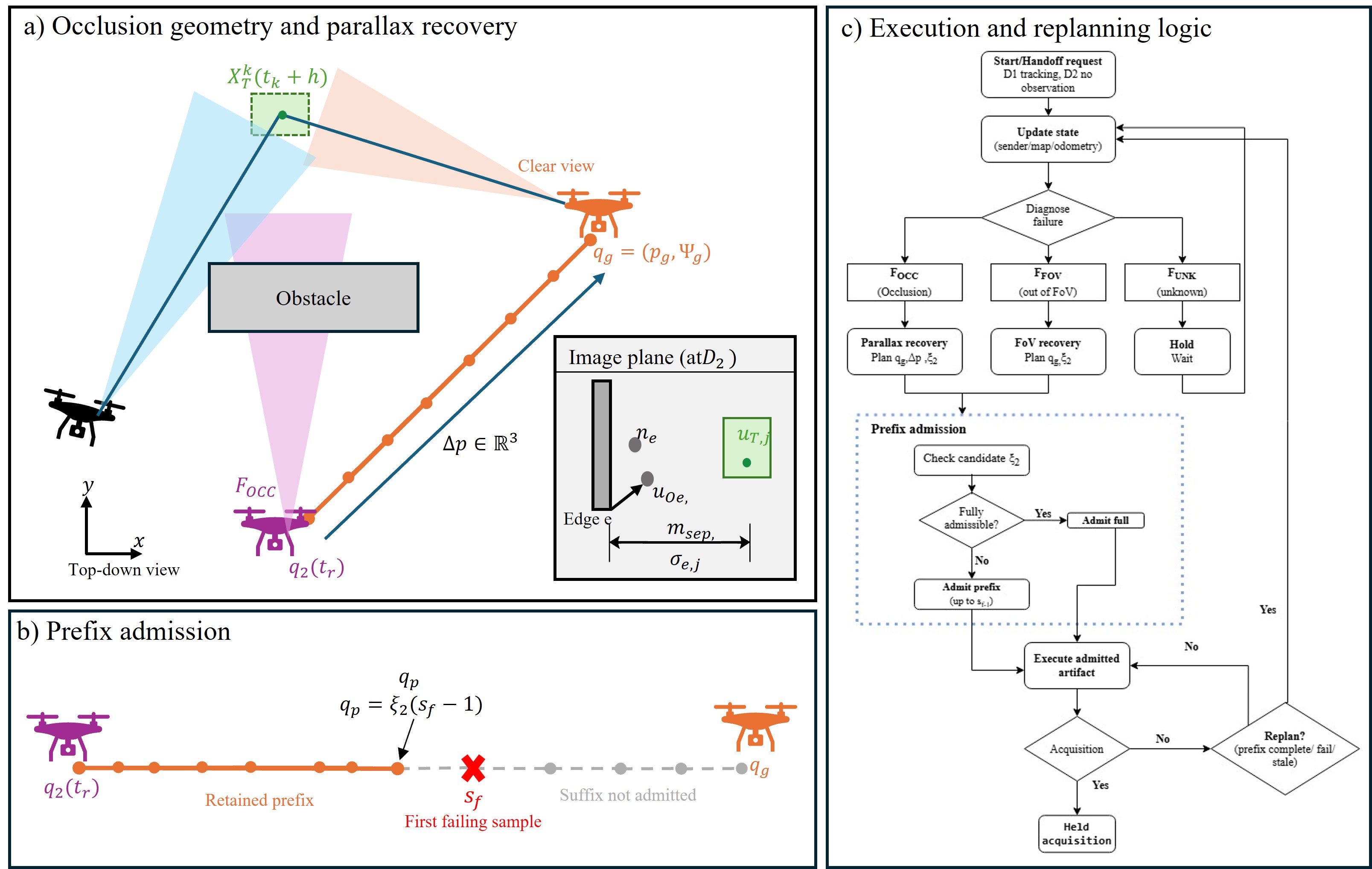}
    \caption{(a) Parallax recovery under $F_{\mathrm{OCC}}$, illustrating the target-support and blocker-edge geometry used to construct a recovery goal $q_g$.
    (b) Sampled admission of the candidate trajectory $\xi_2$, where the retained prefix ends at $q_p=\xi_2(s_{f-1})$ before the first failing sample.
    (c) Runtime execution, held-acquisition checking, and replanning from updated evidence.
\vspace{-2em}}
    \label{fig:method}
\end{figure}

HARP registers sender support, diagnoses viewing conditions, and selects one failure-specific proposal for sampled admissibility checking and execution admission. When full artifact admission fails, HARP may retain a useful nonzero prefix for execution admission. Its endpoint need not satisfy the recovery condition. Updated evidence supports subsequent attempts within a finite search without global completeness or minimum-time optimality. Fig.~\ref{fig:method} provides a geometric interpretation of the parallax recovery, prefix admission, and execution logic used below.

\subsection{Registered Sender Information and Local Mapping}
\label{sec:method_inputs}

HARP uses PATH's bootstrap and fiducial relative-pose interface~\cite{kim2026pathcontinuoustargetsensing}. Writing ${}^{A}\mathbf T_B$ for the transform from $\mathcal F_B$ to $\mathcal F_A$, registration uses navigation and PnP samples matched to the source time within the configured tolerance:
\begin{equation}
{}^{M}\mathbf T_{v^k}
={}^{M}\mathbf T_2(t_k)
\left({}^{v^k}\mathbf T_2(t_k)\right)^{-1}.
\label{eq:method_registration}
\end{equation}
Calibrated camera-to-body extrinsics complete the composition. Positions are rotated and translated, while velocities are rotated. Inputs must be available, valid, unexpired, correctly framed, ordered, and temporally consistent.

$D_1$ supplies component-wise uncertainty parameters
$(\bar{\mathbf e}_0,\bar{\mathbf e}_v,\bar{\mathbf a}_T)$.
For $\Delta_k(\tau)\in[0,H_k]$,
\begin{align}
{}^{v^k}\bar{\mathbf p}_{T,k}(\tau)
&={}^{v^k}\hat{\mathbf p}_{T,k}
+\Delta_k(\tau)\,{}^{v^k}\hat{\mathbf v}_{T,k},\\
\mathbf r_k(\tau)
&=\bar{\mathbf e}_0+\Delta_k(\tau)\bar{\mathbf e}_v
+\tfrac12\Delta_k^2(\tau)\bar{\mathbf a}_T .
\label{eq:method_support_radius}
\end{align}
With ${}^{M}\mathbf T_{v^k}=(\mathbf R_k,\mathbf d_k)$,
\begin{equation}
\begin{aligned}
\mathcal X_T^k(\tau)=\bigl\{&
\mathbf R_k[{}^{v^k}\bar{\mathbf p}_{T,k}(\tau)\\
&+\operatorname{diag}(\mathbf r_k(\tau))\mathbf u]
+\mathbf d_k:\\
&\mathbf u\in[-1,1]^3\bigr\}.
\end{aligned}
\label{eq:method_registered_support}
\end{equation}
The eight transformed vertices preserve the orientation of the registered support. New valid messages replace the active model, and expired models are not extrapolated.

Valid depth hits provide FREE rays and OCCUPIED endpoints, while valid clear-range observations provide only FREE evidence. Invalid observations provide none. Map shifts retain overlapping evidence and initialize new coverage as UNKNOWN. Planning preserves each snapshot's voxel array and timestamp, and accepted map updates during execution request revalidation.

\subsection{Perception-First Diagnosis}
\label{sec:method_diagnosis}

Fresh held acquisition takes priority over diagnosis, which requires a valid registered model and matched camera and map evidence. At synchronized time $t_r$, failed complete-support image readiness yields $F_{\mathrm{FOV}}$. Otherwise, OCCUPIED intersections yield $F_{\mathrm{OCC}}$, including mixed OCCUPIED/UNKNOWN cases. UNKNOWN alone yields $F_{\mathrm{UNK}}$, while a known-FREE view without detection remains waiting for acquisition. Unavailable or rejected inputs are distinct from map UNKNOWN. Model expiration clears current geometry.

For target-camera center $\mathbf c_2(\mathbf q_2)$, the complete viewing volume is
\begin{equation}
\mathcal V_T^k(\mathbf q_2,\tau)
=\bigcup_{\mathbf p\in\mathcal X_T^k(\tau)}
[\mathbf c_2(\mathbf q_2),\mathbf p).
\label{eq:method_viewing_volume}
\end{equation}
Map intersection checks cover this volume with conservative boundary handling and exclude target-endpoint-only contacts.

Future proposals use a finite source-relative horizon set $\mathcal G_k\subseteq[0,H_k]$. For snapshot time $t_s$, each eligible $h\geq t_s-t_k$ is classified independently using geometry at $t_k+h$, with motion budget $h-(t_s-t_k)$. Thus the current diagnosis does not gate every future branch. UNKNOWN-only horizons generate no active exploration proposals, although another future horizon may support FoV or OCC recovery.

\subsection{Failure-Specific Viewpoint Proposals}
\label{sec:method_proposals}

\paragraph{FoV recovery}
At each eligible FoV horizon, HARP first searches over yaw angles at the current position for an endpoint satisfying remaining-time reachability, complete-support $\mathsf{FoVReady}$, and sender-marker observability. If no such endpoint exists, HARP searches a finite translation lattice within a local sphere with inner yaw enumeration and known-FREE endpoint clearance. It returns the first feasible endpoint at the earliest successful eligible $F_{\mathrm{FOV}}$ horizon. Yaw-only body clearance is checked during trajectory admissibility checking.

\paragraph{Occlusion recovery}
OCC recovery uses the projected silhouette of the nearest intersecting occupied component at each eligible $F_{\mathrm{OCC}}$ horizon. Selected boundary edges provide positive-depth anchors and normals toward the free image side. Fig.~\ref{fig:method} (a) illustrates the blocker edge geometry, parallax translation and selected recovery goal used in the OCC proposal.

The standard relationship between camera translation and image motion~\cite{chaumette2006visual} gives the local parallax gradient.
For support vertex $\mathbf p_{T,j}^{C}(h)$, blocker anchor $\mathbf p_{O,e}^{C}$, calibrated pinhole projection $\pi$ in pixel coordinates, and unit normal $\mathbf n_e$,
\begin{align}
\sigma_{e,j}(h)
&=\mathbf n_e^\top[
\pi(\mathbf p_{T,j}^{C}(h))-\pi(\mathbf p_{O,e}^{C})],\\
\mathbf g_{e,j}^{C}(h)
&=[\mathbf J_\pi(\mathbf p_{O,e}^{C})
-\mathbf J_\pi(\mathbf p_{T,j}^{C}(h))]^\top\mathbf n_e,\\
\mathbf g_{e,j}^{M}(h)
&={}^{M}\mathbf R_C\,\mathbf g_{e,j}^{C}(h),
\label{eq:method_parallax_gradient}
\end{align}
where $\mathbf J_\pi=\partial\pi/\partial\mathbf p^C$.
Here $\mathbf p_{T,j}^{C}(h)$ is evaluated at $t_k+h$ in the proposal camera frame.
With orientation fixed for linearization, the proposal solves
\begin{equation}
\begin{aligned}
\min_{\Delta\mathbf p\in\mathbb R^3}\quad&
\tfrac12\|\Delta\mathbf p\|_2^2\\
\mathrm{s.t.}\quad&
\sigma_{e,j}(h)+(\mathbf g_{e,j}^{M}(h))^\top\Delta\mathbf p
\geq m_{\mathrm{sep}},\quad j=1,\ldots,8 .
\end{aligned}
\label{eq:method_parallax_proposal}
\end{equation}
The pixel margin $m_{\mathrm{sep}}$ sets the required separation. Proposals exceeding $r_{\mathrm{trust}}$ or requiring no displacement to satisfy all edge constraints are rejected. No displacement floor or certified linearization-error bound is imposed.

To check the linearized proposal, HARP enumerates yaw at the proposed translation, reprojects the selected 3-D edge and support vertices, and recomputes the normal at $\mathbf q_g$. Exact separation requires, within numerical tolerance,
\begin{equation}
\min_j\mathbf n_e(\mathbf q_g)^\top
[\mathbf u_{T,j}(\mathbf q_g,h)-\mathbf u_{O,e}(\mathbf q_g)]
\geq m_{\mathrm{sep}}.
\label{eq:method_exact_edge}
\vspace{-3pt}
\end{equation}
Complete map visibility then accounts for other blockers.

A \textsc{FullEndpoint} satisfies reachability, known-FREE clearance, target image readiness, marker observability, exact edge separation, and complete known-FREE target visibility at the endpoint. When full admission is prevented by timing, clearance, or UNKNOWN future visibility, HARP may retain a \textsc{PrefixCapableGeometricTarget} if image readiness, basic marker projection and facing geometry, exact edge separation, and no OCCUPIED marker ray contradiction still hold. OCCUPIED target view intersections reject the goal, whereas UNKNOWN prevents full endpoint admission. 

Full OCC candidates are ranked by displacement, then earlier horizon and edge index. If none is available, prefix capable candidates are ranked by greatest remaining model validity at publication, then displacement and edge index. OCC candidates take priority over FoV candidates. Only the selected candidate is passed to trajectory admissibility checking, with no same snapshot fallback after rejection. 



\subsection{Trajectory Construction and Sampled Admissibility}
\label{sec:method_certification}

A selected goal $q_g=(p_g,\psi_g)$ defines a cubic smoothstep trajectory from the current receiver state. With $a(u)=3u^2-2u^3$,
\begin{align}
p_2(s) &= p_0 + a(s/T_\xi)(p_g-p_0),\\
\psi_2(s) &= \operatorname{wrap}
\left(\psi_0+a(s/T_\xi)\Delta\psi\right),
\end{align}
where the duration for nonzero motion is
\begin{equation}
T_\xi =
\max\left\{
\frac{1.5\|p_g-p_0\|_2}{v_{\max}},
\frac{1.5|\Delta\psi|}{\dot{\psi}_{\max}}
\right\}.
\label{eq:traj-duration}
\end{equation}
The same duration is used for proposal reachability. The trajectory is sampled according to the configured time, distance, and yaw increments, including both endpoints.

At planning time $t_p$, each sample $s_i$ queries target and sender geometry at $t_p+s_i$ and must satisfy
\begin{equation}
(t_p-t_k)+s_i \le h \le H_k.
\label{eq:horizon-validity}
\end{equation}
Each sampled receiver pose must have known-FREE clearance, preserve the protected sender--target viewing volume, and maintain sender-marker
observability. At the initial sample, a fresh valid fiducial observation may establish current marker observability when predicted geometry is UNKNOWN and no OCCUPIED contradiction is present.

Full-artifact admission additionally rechecks $\mathrm{FoVReady}$ at the endpoint. OCC recovery also requires exact edge separation and complete known-FREE target visibility. These are endpoint requirements rather than path invariants. If the first failing sample has index $f>1$, HARP may retain the nonzero prefix ending at $s_{f-1}$, excluding the failing and subsequent samples. Let $q_p=\xi_2(s_{f-1})$ denote the retained prefix endpoint. Fig.~\ref{fig:method} (b) illustrates this sample prefix construction. The retained prefix is admissible with respect to the sampled nominal reference and need not satisfy the final recovery condition. These finite-sample checks do not provide a continuous-time guarantee for the executed closed-loop trajectory.

FASTER maintains a backup trajectory in known free space planning through unknown space~\cite{tordesillas2022faster}. In contrast, HARP retains a sampled admissible prefix of the selected recovery motion when the complete artifact cannot be admitted.

\subsection{Receding Execution and Acquisition}
\label{sec:method_execution}
Before execution, HARP checks artifact freshness, identity, start state consistency, and temporal validity. A change in the active sender information requires successful revalidation of the stored nominal samples when the recovery mode supports it. At authorization time, $t_a$, the retained duration $T_{\mathrm {auth}}$ must remain within the applicable target and sender validity intervals and the acquisition deadline $T_A$. 

The controller then executes a smoothstep reference to the retained endpoint with bounded feedback. Before motion commands, HARP checks current state receiver clearance, marker observability, and protection of the sender target sightline. New sender information or accepted map updates trigger revalidation or replanning. Fig.~\ref{fig:method} (c) summarizes the resulting execution, revalidation, and replanning logic. FoV motions and parallax prefix may reuse stored nominal samples, while continuation of a full parallax motion requires regeneration of the blocker dependent geometry. 

Planning time admissibility and execution time guarding are separate. The admissibility check is performed on finite nominal samples, whereas the reference generated at authorization may contain different intermediate poses and time indexed predictions. Runtime guards and revalidation therefore do not guarantee that every executed intermediate pose satisfies all planning constraints.

Recovery terminates upon fresh held acquisition, including during motion. Otherwise, failed continuation, stale or expired evidence, prefix completion, or arrival without acquisition returns $D_2$ to hold and replanning. If no admissible motion is available, HARP withholds additional recovery motion and continues sensing until new evidence supports replanning. Invalid timing, odometry, or tracking state may abort active recovery, while model expiry only removes motion authority.

%% file: Sections/simulation.tex
\section{Simulation Experiments}
\label{sec:sim}

\subsection{Platform and Evaluation Criteria}

\begin{table}[t]
\caption{Primary R0 settings.\vspace{-1em}}
\label{tab:harp-settings}
\centering\footnotesize
\begin{tabularx}{\columnwidth}{@{}lY@{}}
\toprule
Item & Setting\\
\midrule
Sender bounds & $H_k=2$\,s; $e_0=0.10$\,m; $e_v=0.05$\,m/s per axis; $a_T=0$; at most 10\,Hz\\
Association & Pose-matching tolerance 0.08\,s\\
Query entries & $0,0.5,0.75,1,1.25,1.5,1.75,2$\,s\\
FoV/OCC search & FoV: 0.25\,m lattice, 1\,m radius; OCC: 0.75\,m trust radius, 2\,px edge margin; both: $2^\circ$ yaw grid\\
Motion/checks & 0.35\,m/s, 0.20\,rad/s; nominal sampling limits 0.10\,s, 0.05\,m, 0.02\,rad; 0.125\,m body half-extents\\
Depth/local map & $160{\times}120$, $f_x=f_y=120$\,px; depth 0.2--5\,m; map voxels 0.10\,m\\
\bottomrule
\end{tabularx}
\vspace{-2em}
\end{table}

We begin by evaluating sender-guided first acquisition in a ROS-based simulation environment adapted from Eva-Tracker~\cite{lin2026evatracker}. The simulator uses a simplified UAV model with a 30\,Hz world step, static voxel geometry, synthetic depth, and fiducial corners for PnP registration. HARP receives sender predictions and uncertainty bounds, matched sender motion, receiver navigation, and a depth-derived local map. Target truth and complete scene geometry are used only to generate simulated observations and for offline evaluation.

The sender observation proxy applies centroid FoV and range gates. Continued target observation by $D_1$ is an operating assumption. Receiver detections are generated independently by projecting the physical target box and raycasting scene geometry. The visible fraction is the number of visible target pixels divided by the number of physical-target pixels projected within the image and valid depth range, excluding pixels outside either range. Held acquisition requires three consecutive valid bounding-box observations, each with at least 20 visible pixels and a visible fraction $\geq 0.60$, with increasing timestamps and age $\leq 0.5$\,s.

The episode timeout is 120\,s from the first accepted registered model. Acquisition latency is measured from diagnosis to held acquisition and is not assigned to timeout runs. For R0, recovery path length is accumulated from the first EXECUTING event to held acquisition. Comparison studies use their documented diagnosis-to-termination intervals, and the populations are reported separately. Table~\ref{tab:harp-settings}  lists the primary R0 settings.

\subsection{Scenarios and Comparison Design}

F1 tests FoV recovery through attitude change, while F2 requires translation under the configured endpoint policy. O1 introduces a local blocker with a stationary target, O2 combines occlusion with target motion, and C3D evaluates a moving target among cluttered three-dimensional structures. Eligible first-acquisition trials require zero prior valid $D_2$ target observations. Fig.~\ref{fig:entire_sim} shows representative R0 scene geometry and recorded recovery motion for F1, F2, O1, O2, and C3D.

R0 is the primary configuration in Table~\ref{tab:harp-r0}. Three additional arms evaluate configuration sensitivity. R1 increases the target acceleration support to $\bar a_T=0.10$\,m/s$^2$, R2 coarsens the OCC yaw grid from $2^\circ$ to $5^\circ$, and R3 applies both changes. All other settings remain those of R0 unless stated otherwise. R1 and R3 therefore probe sensitivity to faster support growth, while R2 and R3 probe sensitivity to coarser occlusion search.

The evaluation comprises 160 trials across R0--R3. R0 includes one F1 regression run and 16 trials each for F2, O1, O2, and C3D. The canonical F1--O2 results are reported separately from the C3D stress population.

Internal ablations modify one HARP component at a time while retaining common sender-model settings, navigation and map inputs, dynamics, acquisition criteria, and timeout. No-active disables recovery execution. Yaw-only removes translation while retaining endpoint and timing checks. No-prefix admits only complete motion artifacts. Centroid-only uses a point target representation for recovery while retaining complete support for sender protection. The eight-run pilot is a targeted internal ablation study with one execution per setting and no retries, intended to expose mechanism-level differences rather than provide a statistical comparison.  

The comparison study further includes a four-attempt O1 Full/No-prefix repetition, and a prospective O1 geometry matrix. Captured-input replay is used separately when identical-input decision comparison is required.

%% file: Sections/Results_v2.tex
\section{Results}
\label{sec:results}

\begin{figure*}[t]

    \centering
    \includegraphics[width=0.98\linewidth]{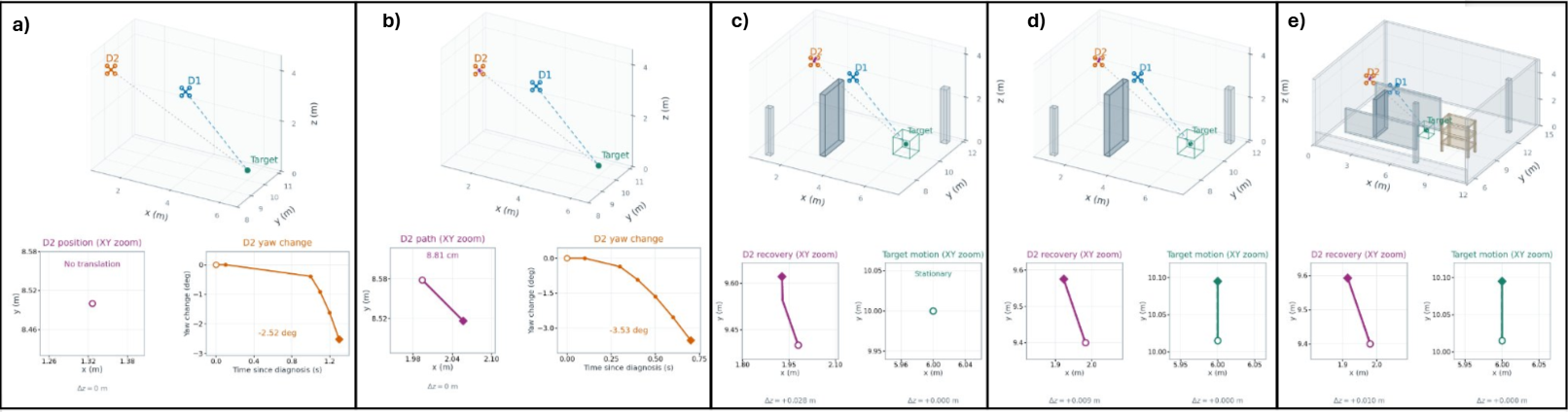}
     \vspace{-8pt}
    \caption{\textbf{Representative R0 recovery scenarios and recorded motion.} (a) F1, (b) F2, (c) O1, (d) O2, and (e) C3D. Lower panels show the recorded $D_2$ and target motion from diagnosis to held acquisition.\vspace{-2em}}
    \label{fig:entire_sim}
\end{figure*}

\subsection{First Acquisition and Receiver Viewpoint}

\begin{figure}[h!]
    \centering
    \includegraphics[width=0.9\linewidth]{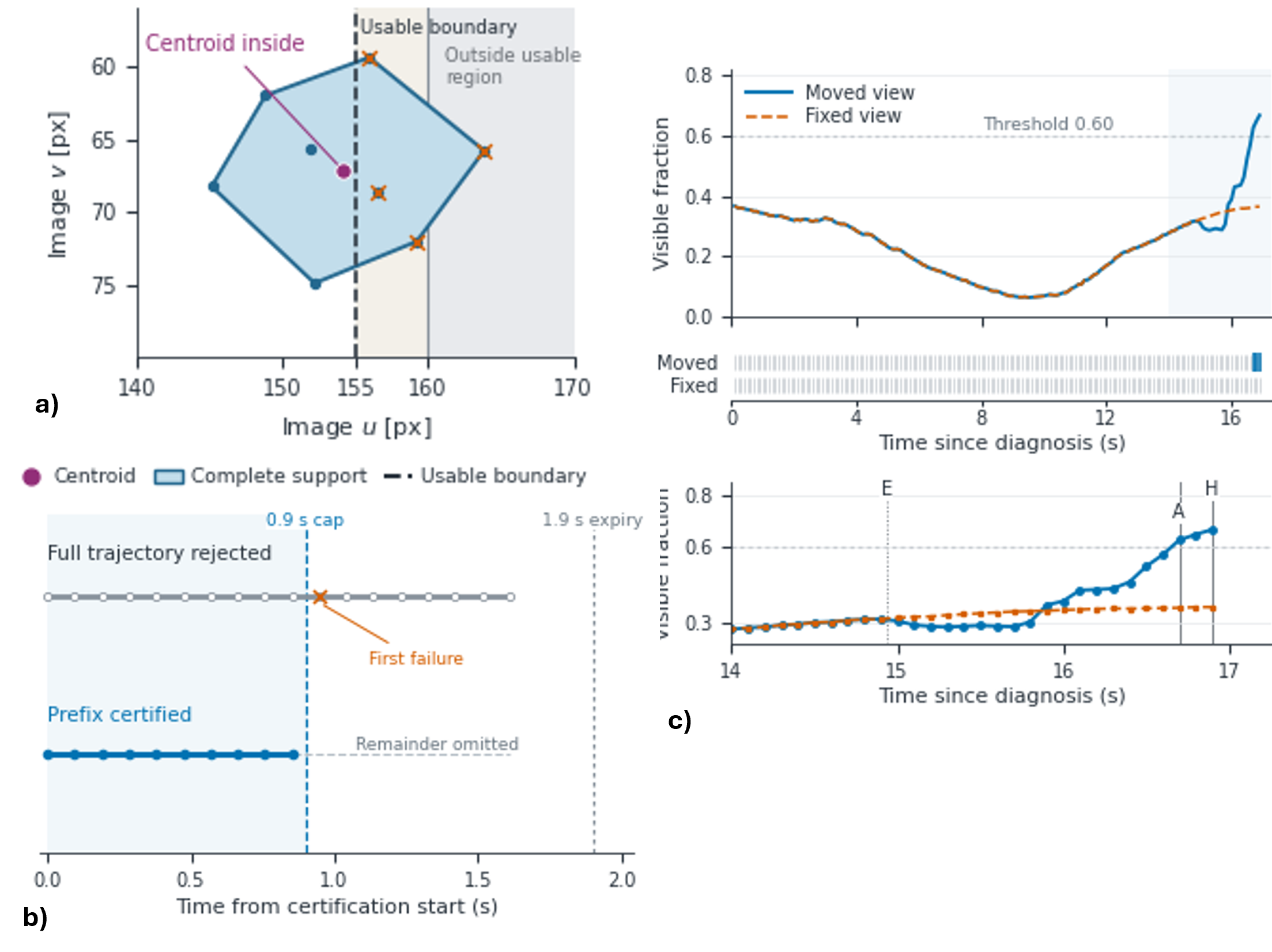}
     \vspace{-10pt}
    \caption{Controlled evidence for HARP recovery decisions.
    (a) \textbf{F2 FoV recovery:} centroid-only checking admits the full artifact, whereas complete-support checking retains only a prefix for the same candidate.
    (b) \textbf{O1 occlusion recovery:} same-input replay shows that prefix admission retains motion when full-artifact admission fails.
    (c) \textbf{O2 moving-target occlusion recovery:} the moved viewpoint becomes acquisition-eligible, whereas the fixed diagnosis pose remains ineligible.\vspace{-6mm}}

    \label{fig:combine}
\end{figure}

HARP achieved held acquisition in all 49 canonical R0 trials and 15/16 C3D stress trials (Table~\ref{tab:harp-r0}). F1 reached held acquisition after 1.3\,s with zero translation and $-2.517^\circ$ yaw. F2 consistently held acquisition in 0.7\,s, while O1 had a median latency of 3.1\,s and a median path length of 0.2225\,m. These results show sender-guided first acquisition under the simulated observation criterion used in the evaluation. C3D is reported separately as a stress population, and latency and path statistics are computed over acquired runs only.


O2 acquired in every R0 trial. Its 62.8 s latency tail involved only 0.0962 m of receiver travel, indicating that long recovery can arise primarily from hold and replanning intervals rather than receiver displacement. The recorded sequence does not isolate a single source of this delay. The higher median latency in O2 than in C3D was concentrated in the interval from diagnosis to the first recorded EXECUTING state. Among acquired runs, the median interval from that state to held acquisition was similar in both scenarios.


Moved-camera replay of representative O2 run 0008 reproduced all 169 recorded samples (Fig.~\ref{fig:combine}(c)). When D2 was instead fixed at its diagnosis pose, none of the 169 samples became acquisition-eligible, with a maximum visible fraction of 0.366617 below the 0.60 threshold. First recorded EXECUTING, eligibility, and held acquisition occurred at 14.933, 16.7, and 16.9\,s after diagnosis. With the target, D1, scene, and timestamps unchanged, this intervention isolates the receiver viewpoint change as necessary for acquisition along this recorded episode. It does not evaluate alternative closed-loop passive policies.


\begin{table}[t]
\caption{R0 held acquisition results\vspace{-1em}}
\label{tab:harp-r0}
\centering\footnotesize
\begin{tabular}{@{}lclc@{}}
\toprule
Scenario & Held/trials & Latency: median [Q1,Q3] s & Path m\\
\midrule
F2 & 16/16 & 0.7 [0.700, 0.700] & 0.0880\\
O1 & 16/16 & 3.1 [3.000, 3.300] & 0.2225\\
O2 & 16/16 & 16.8 [16.700, 16.975] & 0.1755\\
C3D & 15/16 & 7.0 [6.900, 8.800] & 0.0790\\
\bottomrule
\end{tabular}
\par\smallskip\raggedright\footnotesize
Latency ranges: F2 0.7, O1 2.8--3.4, O2 16.4--62.8, and C3D 6.5--10.2\,s. Path entries are medians.\vspace{-2em}
\end{table}

\subsection{Receiver Motion and Support-Dependent Admission}

In the targeted ablation pilot (Table~\ref{tab:ablation}), Full acquired in F1, F2, and O1, while No-active in F1 and Yaw-only in F2 timed out. Replay of the F2 Yaw-only case identified a geometrically feasible rotation that exceeded the implemented yaw-rate and timing envelope. The result therefore shows that the configured translation-and-yaw policy admitted a recovery that the implemented Yaw-only variant did not, rather than establishing geometric impossibility of yaw-only recovery. No-prefix also timed out in O1, whose mechanism is examined below. Centroid-only acquired in both tested scenes.

For the same F2 centroid-generated candidate, centroid checking admitted a 1.515229\,s full artifact, whereas complete-support checking identified an arrival-FoV failure and retained a 1.420527\,s prefix (Fig.~\ref{fig:combine}(a)). Complete-support reasoning therefore changed the admissible action extent for an otherwise identical candidate. This directly demonstrates support-dependent admission for the tested candidate. The comparison concerns admission behavior rather than a general acquisition advantage over centroid reasoning.

\newcommand{\timeout}{\textsc{Timeout}}
\begin{table}[h!]
\caption{Internal ablation Pilot and O1 prefix repetitions\vspace{-1em}}
\label{tab:ablation}
\centering\scriptsize
\begin{tabular}{@{}lllr@{}}
\toprule
Scenario & Mode & Outcome/time & Path (m)\\
\midrule
\multicolumn{4}{@{}l}{\textit{Eight-run pilot}} \\
F1 & Full / No-active & 1.4 s / \timeout & 0 / 0\\
F2 & Full / Yaw-only & 0.7 s / \timeout & .0880 / 0\\
F2 & Centroid-only & 0.7 s & .1325\\
O1 & Full / No-prefix & 3.4 s / \timeout & .2084 / 0\\
O1 & Centroid-only & 4.5 s & .2472\\
\midrule
\multicolumn{4}{@{}l}{\textit{Four-attempt O1 repetition}} \\
\addlinespace[2pt]
O1 att. 1 & Full & 3.0 s & .2148\\
O1 att. 2 & No-prefix & \timeout & 0\\
O1 att. 3 & No-prefix & \timeout & 0\\
O1 att. 4 & Full & 3.2 s & .2164\\
\bottomrule
\end{tabular}
\vspace{-20pt}
\end{table}

\subsection{Prefix-Enabled Progress and Geometry Sensitivity}

Same-input O1 replay isolated the effect of prefix permission on a selected trajectory that failed full-artifact admission (Fig.~\ref{fig:combine}(b)). With source stamp $t_k=7.9$\,s, planning time $t_p=8.0$\,s, selected horizon $h=1$\,s, and lifetime $H_k=2$\,s, the remaining budgets were
\begin{equation}
\begin{aligned}
h-(t_p-t_k)&=0.9\,\mathrm{s},\\
H_k-(t_p-t_k)&=1.9\,\mathrm{s}.
\end{aligned}
\label{eq:harp-prefix-budget}
\end{equation}
The full 1.609506\,s trajectory fit the remaining lifetime but exceeded the selected-horizon cap in Eq.~\eqref{eq:harp-prefix-budget}. Its first failing sample occurred at 0.946768\,s, leaving an admissible prefix ending at 0.852091\,s. With prefix permission, the trajectory checker retained this portion of the selected motion. Without prefix permission, the selected trajectory produced no admitted motion artifact. The replay therefore isolates the effect of prefix permission on admission for the same input.

Live O1 repetitions produced acquisition times of 3.0 s/TO/TO/3.2 s in the order Full/No-prefix/No-prefix/Full. Both No-prefix runs remained stationary. Although the maps and poses were realized separately for each execution, the reversed ordering under common settings provides live evidence consistent with the identical-input replay.

The prospective geometry study extended this comparison across variations in staging offset and blocker width (Table~\ref{tab:harp-geometry}). Two conditions were excluded because $D_2$ had already observed the target before recovery. Full acquired in three of four eligible conditions, whereas No-prefix acquired in none. G01 and G02 reproduced the selected-horizon limitation observed in the replay. In G06, No-prefix failed complete arrival visibility at $h=2$ s, while Full executed a 0.389330\,s prefix and acquired, extending the observed effect beyond the 1\,s horizon case.

G04 exposed a different failure boundary. Proposal generation and selection failed before primary trajectory checking in both modes because the finite generator produced no admissible selected candidate. Prefix permission therefore could not provide progress. This case shows that prefix admission can improve progress after a candidate is generated, but does not overcome limitations of the upstream finite recovery search.

\begin{table}[t]
\caption{All O1 Geometry study\vspace{-1em}}
\label{tab:harp-geometry}
\centering\footnotesize
\begin{tabular}{@{}lrrll@{}}
\toprule
ID & $\Delta Y$ m & Width m & Full & No-prefix\\
\midrule
G01 & 0.00 & 0.15 & 4.1\,s & TO\\
G02 & 0.00 & 0.25 & 3.4\,s & TO\\
G03 & $-0.20$ & 0.15 & Prior obs. & Skipped\\
G04 & $-0.20$ & 0.25 & TO & TO\\
G05 & $+0.20$ & 0.15 & Prior obs. & Skipped\\
G06 & $+0.20$ & 0.25 & 0.4\,s & TO\\
\bottomrule
\end{tabular}
\end{table}

\subsection{Configuration Sensitivity}

\begin{table}[t]
\caption{Configuration sensitivity\vspace{-1em}}
\label{tab:harp-sensitivity}
\centering\footnotesize
\begin{tabular}{@{}lcccc@{}}
\toprule
Arm / change from R0 & F2 & O1 & O2 & C3D\\
\midrule
R1: $a_T=0.10$\,m/s$^2$ & 0/5 & --- & 5/5 & 4/5\\
R2: $5^\circ$ OCC yaw grid & 15/15 & 15/15 & 14/15 & 13/15\\
R3: both changes & 0/5 & 4/5 & 5/5 & 4/5\\
\bottomrule
\end{tabular}
\vspace{-1em}
\end{table}

R2 retained acquisition throughout F2 and O1, with occasional O2 and C3D timeouts, whereas R1 and R3 failed throughout F2 but mostly acquired in O2 and C3D (Table~\ref{tab:harp-sensitivity}). These results show scenario-dependent sensitivity to target-support growth and occlusion-search resolution. Because the arms used unequal screening and confirmation allocations and individual rejection causes were not isolated, they characterize sensitivity within the tested configurations rather than general robustness to unmodeled uncertainty.


\subsection{Constraint Evidence and Computational Cost}

No violations were observed in the recorded checks examined for receiver clearance, sender fiducial observability, protected-sightline preservation, and model validity in the pilot, O1 repetition, and geometry studies. These findings are limited to the sampled checks and do not establish continuous-time safety.

In the representative C3D run, target-centroid containment held for 5/8 examined selected-arrival queries, with a maximum miss of 0.156769\,m. The three misses occurred at future times beyond the corresponding prefix endpoints and used prediction models different from those active during runtime validation. They therefore do not establish containment violations during execution. A complete execution-wide containment analysis could not be reconstructed from the available time-indexed transforms, support vertices, and authorization mappings.


OCC generation time varied with scene geometry, with medians of about 143\,ms in the two Full O1 repetitions and 456--467\,ms in G04, while one frozen O2 record reached 4.11\,s. We therefore treat HARP as an event-driven recovery planner rather than claim a fixed-rate 10\,Hz planning loop.

Together, the experiments support three aspects of HARP. The canonical R0 acquisitions and O2 receiver-viewpoint intervention demonstrate sender-guided first acquisition in the tested FoV and occlusion conditions, complemented by the bounded C3D stress results. The F2 same-candidate analysis demonstrates support-dependent admission, while the motion comparison shows the role of the configured dynamic and temporal endpoint policy. The O1 replay isolates the effect of prefix permission on admission, while the repetitions and geometry study show live progress with prefix permission across the tested configurations.

\subsection{Flight Validation}
\label{sec:field-results}

\begin{figure}[t]
    \centering
    \includegraphics[width=0.8\linewidth]{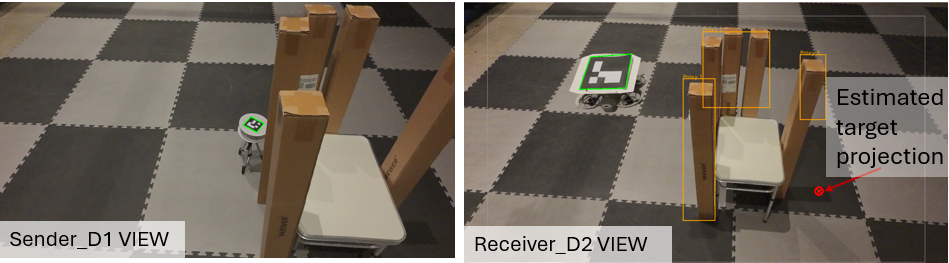}
     \vspace{-6pt}
    \caption{Observations in the two-UAV setup.
    Left: the sender observes the target marker beside the
    obstacles. Right: the receiver observes the sender-mounted
    marker, while the target is not directly visible.\vspace{-1em}}
    \label{fig:hardware-views}
\end{figure}

We evaluated HARP using two UAVs under field-of-view loss and environmental occlusion while the sender maintained target observation. The evaluation considered three recovery cases: yaw-only FoV recovery, lateral recovery from static occlusion, and recovery of a moving target behind a blocker. Figure~\ref{fig:hardware-views} shows the sensing configuration: the sender maintains a target observation, while the receiver must obtain a usable view despite intervening obstacles.

\begin{figure}
    \centering
    \includegraphics[width=\linewidth]{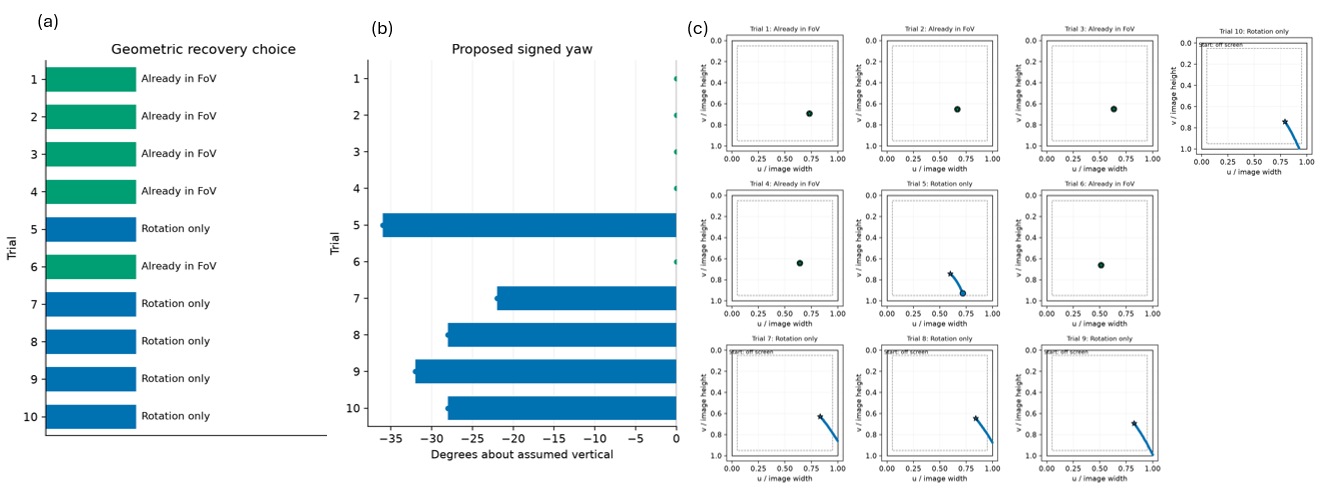}
    \vspace{-17pt}
    \caption{Receiver FoV analysis across ten cases: (a) geometric recovery classification, (b) proposed receiver yaw, and (c) predicted target-image paths
\vspace{-3em}}
    \label{fig:hw-test1}
\end{figure}

For FoV recovery, Figure~\ref{fig:hw-test1} shows rotation-only proposals in five of ten flights, with yaw magnitudes of approximately $22$--$36^\circ$. In the remaining cases, the estimated target support already lay inside the receiver FoV. These results show how the recovery formulation distinguishes cases that require camera reorientation from those that require further visibility assessment. In particular, an inside-FoV target can remain unavailable for acquisition because of occlusion. 

\begin{figure}
\vspace{-1em}
    \centering
    \includegraphics[width=0.95\linewidth]{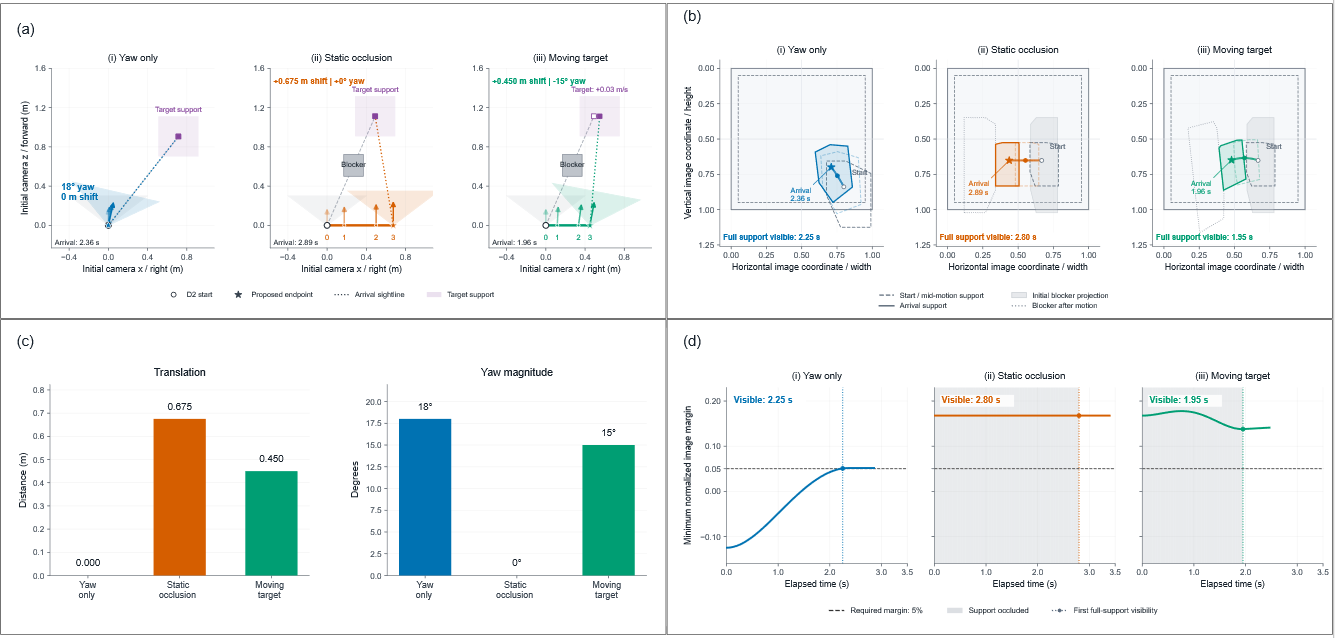}
    \vspace{-10pt}
    \caption{Recovery analysis for three scenarios: (a) receiver paths and target geometry, (b) projected target uncertainty regions, (c) proposed translation and yaw, and (d) minimum image-edge margin, with shaded intervals indicating occlusion.}
    \label{fig:hw-test2}
\end{figure}

Figure~\ref{fig:hw-test2} compares the receiver motions proposed for three recovery cases. The yaw-only case requires an $18^\circ$ rotation without translation. Static occlusion requires a $0.675$\,m lateral displacement without additional yaw, while the moving-target case combines a $0.450$\,m lateral displacement with a $ 15^\circ$ yaw adjustment. These viewpoint changes produce the target-image paths shown in panel~(b). The full target support first satisfies both the required image margin and the occlusion-clearance condition at approximately $2.25$\,s, $2.80$\,s, and $1.95$\,s, respectively. The corresponding proposed arrival times are $2.36$\,s, $2.89$\,s, and $1.96$\,s. Panel~(d) explains why FoV inclusion alone is insufficient: in both occlusion cases, the image margin exceeds the required $5\%$ threshold while part of the target support remains blocked. Lateral motion changes the viewing direction around the blocker, while yaw adjusts the target projection within the image. These geometric predictions support HARP's recovery rationale: selecting motion based on the cause of acquisition failure and evaluating visibility across the entire target support.

Across the three model-based cases, the median time to satisfy the geometric visibility conditions was $2.25$\,s, and the median proposed translation distance was $0.450$\,m. All three cases satisfied these conditions in approximately $2.80$\,s.

%% file: Sections/Conclusion.tex
\section{Conclusion}


This work presented HARP, a receiver-side recovery method for sender-guided first acquisition during multi-UAV handoff. HARP uses information from the sending UAV to identify field-of-view and occlusion failures, generate recovery viewpoints, and evaluate receiver motion under target uncertainty and handoff constraints. When a complete trajectory cannot be admitted, HARP can execute a sampled admissible nonzero prefix and replan using updated observations. Closed-loop simulations and controlled replays demonstrated sender-guided first acquisition and isolated the effects of receiver viewpoint change, target support reasoning, and prefix permission. Field evaluation further validated the corresponding FoV and occlusion recovery geometry in a physical two-UAV setting. Future work will improve planning efficiency, strengthen continuous-time safety guarantees, and extend evaluation to larger, more dynamic environments with uncertain communication, sensing, localization, and target motion during deployment.


%% file: ref.bib
@inproceedings{jeon2019online,
  author = {Jeon, Boseong Felipe and Kim, H. Jin},
  title = {{Online Trajectory Generation of a MAV for Chasing a Moving Target in 3D Dense Environments}},
  year = {2019},
  booktitle = {2019 IEEE/RSJ International Conference on Intelligent Robots and Systems (IROS)},
  pages = {1115--1121}
}

@inproceedings{jeon2020integrated,
  author = {Jeon, Boseong and Lee, Yunwoo and Kim, H. Jin},
  title = {{Integrated Motion Planner for Real-time Aerial Videography with a Drone in a Dense Environment}},
  year = {2020},
  booktitle = {2020 IEEE International Conference on Robotics and Automation (ICRA)},
  pages = {1243--1249}

}

@inproceedings{wang2021visibility,
  author = {Wang, Qianhao and Gao, Yuman and Ji, Jialin and Xu, Chao and Gao, Fei},
  title = {{Visibility-aware Trajectory Optimization with Application to Aerial Tracking}},
  year = {2021},
  booktitle = {2021 IEEE/RSJ International Conference on Intelligent Robots and Systems (IROS)},
  pages = {5249--5256}
}

@article{lee2025qpchaser,
  author = {Lee, Yunwoo and Park, Jungwon and Jung, Seungwoo and Jeon, Boseong and Oh, Dahyun and Kim, H. Jin},
  title = {{QP Chaser: Polynomial Trajectory Generation for Autonomous Aerial Tracking}},
  year = {2025},
  journal = {IEEE Transactions on Automation Science and Engineering},
  volume = {22},
  pages = {24177--24194}
}

@article{lee2025dmvc,
  author = {Lee, Yunwoo and Park, Jungwon and Kim, H. Jin},
  title = {{DMVC-Tracker: Distributed Multi-Agent Trajectory Planning for Target Tracking Using Dynamic Buffered Voronoi and Inter-Visibility Cells}},
  journal = {IEEE Robotics and Automation Letters},
  year = {2025},
  volume = {10},
  number = {5},
  pages = {4842--4849}
}

@article{chaumette2006visual,
  author = {Chaumette, Fran{\c{c}}ois and Hutchinson, Seth},
  title = {{Visual servo control. I. Basic approaches}},
  year = {2006},
  journal = {IEEE Robotics \& Automation Magazine},
  volume = {13},
  number = {4},
  pages = {82--90}
}

@article{tordesillas2022faster,
  author = {Tordesillas, Jes{\'u}s and Lopez, Brett T. and Everett, Michael and How, Jonathan P.},
  title = {{FASTER: Fast and Safe Trajectory Planner for Navigation in Unknown Environments}},
  year = {2022},
  journal = {IEEE Transactions on Robotics},
  volume = {38},
  number = {2},
  pages = {922--938}
}

@misc{lin2026evatracker,
  author = {Lin, Yue and Liu, Yang and Wang, Dong and Lu, Huchuan},
  title = {{Eva-Tracker: ESDF-update-free, Visibility-aware Planning with Target Reacquisition for Robust Aerial Tracking}},
  year = {2026},
  note = {arXiv:2602.12549}
}

@article{yin2026visibility,
  title={Visibility-Aware Cooperative Tracking With Decentralized LiDAR-Based Aerial Swarms},
  author={Yin, Longji and Ren, Yunfan and Zhu, Fangcheng and Shi, Liuyu and Kong, Fanze and Tang, Benxu and Liu, Wenyi and Lyu, Ximin and Zhang, Fu},
  journal={IEEE Transactions on Robotics},
  year={2026},
  publisher={IEEE}
}

@inproceedings{wu2025learning,
  title={Learning occlusion-robust vision transformers for real-time UAV tracking},
  author={Wu, You and Wang, Xucheng and Yang, Xiangyang and Liu, Mengyuan and Zeng, Dan and Ye, Hengzhou and Li, Shuiwang},
  booktitle={2025 IEEE/CVF Conference on Computer Vision and Pattern Recognition (CVPR)},
  pages={17103--17113},
  year={2025},
  organization={IEEE}
}

@article{xue2024handling,
  title={Handling occlusion in uav visual tracking with query-guided redetection},
  author={Xue, Yuanliang and Shen, Tao and Jin, Guodong and Tan, Lining and Wang, Nian and Wang, Lianfeng and Gao, Jing},
  journal={IEEE Transactions on Instrumentation and Measurement},
  volume={73},
  pages={1--17},
  year={2024},
  publisher={IEEE}
}

@INPROCEEDINGS{ryou2026Real,
  author={Ryou, Gilhyun and Beyer, Lukas Lao and Karaman, Sertac},
  booktitle={2025 IEEE International Conference on Robotics and Automation (ICRA)}, 
  title={Real-Time Sampling-based Online Planning for Drone Interception}, 
  year={2025},
  volume={},
  number={},
  pages={2801-2807},
  doi={10.1109/ICRA55743.2025.11128529}}

@inproceedings{kim2025continuous,
  title={Continuous Marine Monitoring via Autonomous UAV Handoff},
  author={Kim, Heegyeong and James, Alice and Seth, Avishkar and Kuantama, Endrowednes and Williamson, Jane and Feng, Yimeng and Han, Richard},
  booktitle={Proceedings of the 23rd Annual International Conference on mobile systems, applications and services},
  pages={711--716},
  year={2025}
}

@article{leong2021intelligent,
  title={An intelligent auto-organizing aerial robotic sensor network system for urban surveillance},
  author={Leong, Wai Lun and Martinel, Niki and Huang, Sunan and Micheloni, Christian and Foresti, Gian Luca and Teo, Rodney Swee Huat},
  journal={Journal of Intelligent \& Robotic Systems},
  volume={102},
  number={2},
  pages={33},
  year={2021},
  publisher={Springer}
}

@inproceedings{bhagat2020uav,
  title={UAV target tracking in urban environments using deep reinforcement learning},
  author={Bhagat, Sarthak and Sujit, PB},
  booktitle={2020 International conference on unmanned aircraft systems (ICUAS)},
  pages={694--701},
  year={2020},
  organization={IEEE}
}

@article{feng2025deep,
  title={Deep reinforcement learning for UAV target search and continuous tracking in complex environments with Gaussian process regression and prior policy embedding},
  author={Feng, Zhihui and Na, Xitai and Hai, Shiji and Sun, Qingbin and Shi, Jinshuo},
  journal={Electronics},
  volume={14},
  number={7},
  pages={1330},
  year={2025},
  publisher={MDPI}
}

@inproceedings{brommer2018long,
  title={Long-duration autonomy for small rotorcraft uas including recharging},
  author={Brommer, Christian and Malyuta, Danylo and Hentzen, Daniel and Brockers, Roland},
  booktitle={2018 IEEE/RSJ International Conference on Intelligent Robots and Systems (IROS)},
  pages={7252--7258},
  year={2018},
  organization={IEEE}
}

@article{zhang2025distributed,
  title={A distributed task allocation approach for multi-UAV persistent monitoring in dynamic environments},
  author={Zhang, Changming and Xu, Caiyue and Li, Gang and He, Bin},
  journal={Scientific Reports},
  volume={15},
  number={1},
  pages={6437},
  year={2025},
  publisher={Nature Publishing Group UK London}
}

@misc{kim2026pathcontinuoustargetsensing,
  title  = {{PATH: Continuous Target Sensing among Autonomous Cooperative Drones}},
  author = {Heegyeong Kim and Alice James and Avishkar Seth and Endrowednes Kuantama and Jane Williamson and Yimeng Feng and Richard Han},
  year   = {2026},
  note   = {arXiv:2609.12456}
}

@article{sun2024moving,
  title={Moving target tracking by unmanned aerial vehicle: A survey and taxonomy},
  author={Sun, Nianyi and Zhao, Jin and Shi, Qing and Liu, Chang and Liu, Peng},
  journal={IEEE Transactions on Industrial Informatics},
  volume={20},
  number={5},
  pages={7056--7068},
  year={2024},
  publisher={IEEE}
}

@article{hu2025improved,
  title={An Improved Hierarchical Framework for Adaptive UAV Trajectory Planning in Dynamic Target Tracking},
  author={Hu, Sen and Tian, Zhong and Zhu, Weiyu and Zhang, Bangchu},
  journal={Drones},
  volume={9},
  number={12},
  pages={847},
  year={2025},
  publisher={MDPI}
}

@article{rao2025multi,
  title={Multi-UAV trajectory planning with field-of-view sharing mechanism in cluttered environments: application to target tracking},
  author={Rao, Kai and Yan, Huaicheng and Yang, Penghui and Wang, Meng and Lv, Yunkai},
  journal={Science China Information Sciences},
  volume={68},
  number={5},
  pages={150206},
  year={2025},
  publisher={Springer}
}

@inproceedings{gao2024probabilistic,
  title={Probabilistic visibility-aware trajectory planning for target tracking in cluttered environments},
  author={Gao, Han and Pengying, W and Yao, S and Zhou, Kangjie and Ji, M and Liu, Hangxin and Liu, Chang},
  booktitle={2024 American Control Conference (ACC)},
  pages={594--600},
  year={2024},
  organization={IEEE}
}

@article{wang2023svpto,
  title={SVPTO: Safe visibility-guided perception-aware trajectory optimization for aerial tracking},
  author={Wang, Hanzhang and Zhang, Xuetao and Liu, Yisha and Zhang, Xuebo and Zhuang, Yan},
  journal={IEEE Transactions on Industrial Electronics},
  volume={71},
  number={3},
  pages={2716--2725},
  year={2023},
  publisher={IEEE}
}

@inproceedings{zhou2024aspire,
  title={Aspire: An informative trajectory planner with mutual information approximation for target search and tracking},
  author={Zhou, Kangjie and Wu, Pengying and Su, Yao and Gao, Han and Ma, Ji and Liu, Hangxin and Liu, Chang},
  booktitle={2024 IEEE International Conference on Robotics and Automation (ICRA)},
  pages={4626--4632},
  year={2024},
  organization={IEEE}
}
